\documentclass{article}
\PassOptionsToPackage{numbers, compress}{natbib}
\usepackage[final]{nips_2017}

\usepackage[utf8]{inputenc} 
\usepackage[T1]{fontenc}    
\usepackage{hyperref}       
\usepackage{url}            
\usepackage{booktabs}       
\usepackage{amsfonts}       
\usepackage{nicefrac}       
\usepackage{microtype}      

\usepackage{algorithm}
\usepackage{algorithmic}

\usepackage{graphicx}
\usepackage{amsmath}
\usepackage[font=small,skip=0pt]{caption}
\usepackage{subfigure}
\title{Unsupervised Learning Layers for Video Analysis}

\author{
  Liang Zhao, Yang Wang, Yi Yang, Wei Xu\\
  Baidu Research - Institute of Deep Learning\\
  Sunnyvale, CA 94089\\
  \texttt{\{zhaoliang07, wangyang59, yangyi05, xuwei06\}@baidu.com}\\ 
}

\begin{document}

\maketitle

\begin{abstract}
This paper presents two unsupervised learning layers (UL layers) for label-free
 video analysis: one for fully connected layers, and the other for 
convolutional ones. The proposed UL layers can play two roles: they can be
 the cost function layer for providing global training signal; meanwhile they
 can be added to any regular neural network layers for providing local training
signals and combined with the training signals backpropagated from upper layers
for extracting both slow and fast changing features at layers of 
different depths. Therefore, the UL layers can be used in either pure unsupervised or semi-supervised settings.  
Both a closed-form solution and an online learning algorithm for 
two UL layers are provided.
Experiments with unlabeled synthetic and real-world videos 
demonstrated that the neural networks equipped with UL layers and trained with the 
proposed online learning algorithm can extract shape and motion information
from video sequences of moving objects.  The experiments demonstrated the potential 
applications of UL layers and online learning algorithm to head orientation estimation and moving object localization.
\end{abstract} 

\section{Introduction}
Deep neural networks (DNNs) are powerful and flexible models that can 
extract hierarchical and discriminative features from large amounts of data.
Despite their success in many supervised tasks such as image 
classification \cite{Krizhevsky12, He16}, speech recognition \cite{Yu12} and 
machine translation \cite{Wu16}, 
DNNs are data hungry which limits their applications in many domains where 
abundant annotated data are not available. This motivates us to explore 
almost infinite amount of unlabeled data for obtaining good representations 
that generalizes across tasks.

Learning from unlabeled data, often named as unsupervised learning, can be 
divided into four levels: learning from unlabeled images, from unlabeled 
videos, from virtual environment, or from the real environment directly. In 
this paper, we focus on unsupervised learning from label-free videos.

One of the main challenges of unsupervised learning is to derive good 
training signals. For supervised learning, we use the difference between the 
prediction from a DNN and the annotated ground truth as training signal. 
For unsupervised learning, there are two ways to obtain such signal: one is 
to design auxiliary tasks such as reconstructing the input image 
\cite{Hinton06b, Bengio07, Le12}, predicting the next fewer frames given the 
first fewer frames in a video \cite{Srivastava15, Finn16, Lotter16}, 
reordering the shuffled video frames \cite{Misra16}, generating fake 
images/videos in a generative adversarial network settings (GAN) 
\cite{Goodfellow14}, etc; another is to provide constraints 
\cite{Wiskott02, Higgins16, Lin16, Stewart17} that describe the desired structure of 
the output from a DNN. Our work belongs to the latter one.

Inspired by human visual system that can learn invariant representations and 
structures of objects from temporal experiences \cite{Foldiak91, Wiskott02}, 
we design an objective function that constrains the output from a DNN to be 
temporally consistent meanwhile avoiding degenerated cases.

We evaluate our proposed algorithm in both synthetic and natural video 
settings. 
Experiments with end-to-end training on unlabeled-videos and applications to 
head orientation estimation and moving object localization demonstrated the 
effectiveness of the proposed algorithm.

The contributions of this paper are the following:
\begin{itemize}
\item We design two unsupervised learning layers (UL layers) for label-free
 video analysis: one for fully connected layers, and the other for
convolutional layers. The proposed UL layers can play two roles: they can be
 the cost function layer for providing global training signal; meanwhile they
 can be added to any regular neural network layers for providing local training
signals and be combined with the training signals backpropagated from upper layers
for extracting both slow and fast changing features at layers of 
different depths.

\item Both a closed-form solution and an online learning algorithm for 
the two proposed UL layers are provided. 

\item The UL layers can be applied to any neural network architectures and 
 can be used in either pure unsupervised or semi-supervised settings.
We evaluated the proposed algorithm on both synthetic and real-world videos,
and show that it has potential application to head orientation estimation, and moving object localization.
\end{itemize}

The rest of the paper is organized as follows. We briefly review related work 
in Section 2. The detailed description of the proposed methods is given in
Section 3 followed by experimental results in Section 4 and conclusion in 
Section 5.

\section{Related Work}
Our work is closely related to representation learning, unsupervised 
learning and its application to label-free video analysis. Here, we briefly review some related work in these areas.

\textbf{Representation Learning:}  Learning good data representations (or 
features) is one of the main goals for any machine learning algorithms -- no 
matter whether it is a supervised or unsupervised one. For unsupervised 
learning, we want the learned representations of the data makes it easier to 
build classifiers or other predictors upon them. The question is how to 
evaluate the quality of learned representations. In \cite{Goodfellow08}, 
Goodfellow et~al. proposed a number of empirical tests for directly measuring 
the degree to which learned representations are invariant to different 
transformations.  In \cite{Higgins16}, Higgins et~al. devised a protocol to 
quantify the degree of disentanglement learned by different models. In 
\cite{Bengio13}, Bengio et~al. list some properties/attributes a good 
representation should possess, such as sparsity, distributed among multiple 
explanatory factors, hierarchical organization with more abstract and 
invariant concepts \cite{Hubel62, Wiskott02} higher in the hierarchy, temporal 
and spatial coherence. The listed properties provide us guide to design 
appropriate objective functions for learning such representations.

\textbf{Unsupervised Learning:}  Unsupervised learning of visual 
representations is a broad area with a rich history and a large volume of 
work -- ranging from classical K-means clustering \cite{MacQueen67}, 
dimensionality reduction \cite{Roweis97, Hinton06a}, sparse coding 
\cite{Olshausen96, Lee07}, RBMs(Restricted Boltzmann Machines)\cite{Hinton06b}, 
autoencoders \cite{Bengio07, Le12}, single-layer analysis \cite{Coates11}, to 
more recent work such as VAEs(Variational Auto-Encoders) \cite{Kingma14}, 
GANs(Generative Adversarial Networks) \cite{Goodfellow14}, pixelCNNs \cite{Oord16b}, and pixelRNNs \cite{Oord16a}. The increase of 
computational capabilities \cite{Raina09}  also makes large-scale deep unsupervised learning possible. 

Our work is closely related to unsupervised learning from 
label-free video. Videos provide import temporal constraints for machine 
learning algorithms and they are abundant on websites. Most recent work employ 
video prediction as an auxiliary task for learning useful representations. The 
argument is that if the algorithm can predict the next fewer frames well, it 
has to learn some features related to objects' shape, motion, etc. These 
algorithms usually predict next frames at pixel \cite{Srivastava15}, 
interstate \cite{Vondrick15}, object location \cite{Stewart17} levels using 
language models \cite{Ranzato14}, motion transformation \cite{Finn16}, LSTM 
\cite{Srivastava15}, GAN \cite{Mathieu16}, or probabilistic models 
\cite{Xue16}.

Unlike the above work that relies on auxiliary tasks for unsupervised learning,
our work belongs to constraints-based unsupervised learning. We design an 
objective function that constrains the output from a DNN to be temporally 
consistent meanwhile avoiding degenerated cases. \cite{Stewart17, Wiskott02, 
Berkes05} are the most related work. In \cite{Stewart17}, the authors propose 
to employ laws of physics as constraints for supervising neural network 
training. For tracking an object in free fall, they constrain the outputs from 
the neural network which encodes the object's height should form a parabola;
while for tracking the position of a walking man, the outputs should satisfy 
the constant velocity constraint. Therefore, they have to design different loss functions for different types of object motion. In contrast, our objective function is more general and can be used to analyze videos of any smoothly moving 
objects. In \cite{Wiskott02}, the authors propose a slow feature analysis 
algorithm for learning invariant or slowly varying features and demonstrate
that the learned functions have a good match with complex cell properties. Similar to our closed-form solution method, their objective function also constrains the output should vary as slowly as possible but keep a unit variance. Like our
 work, they also proposed two neural network implementations of the algorithm.
The difference is that we designed two unsupervised learning layers for implementing our online algorithms. The UL layers can be integrated into any neural networks seamlessly.

\section{Methods}
\subsection{Problem Formulation}
In our label-free learning setting, the training set is $D=\{X_1, ..., X_n\}$ of
$n$ training video sequences. Each sequence $X_i=\{\mathbf{x}_{i1}, ..., \mathbf{x}_{im}\}$
consists of $m$ image frames. The goal is to learn an internal representation 
$Y$ of $X$ and a function $f_{\theta}(\mathbf{x}): 
X \rightarrow Y$ such that

\begin{equation}
f_\theta^* =arg \min_{f \in F} E(||\mathbf{y}_t - \mathbf{y}_{t-1}||^2) - \log\det(cov(\mathbf{y}))
\end{equation}

where the optimization is over a predefined class of functions $F$. The first
term in Eq. (1) enforces temporal consistency constraint, and the second
term is used for avoiding degenerate case where $\mathbf{y}_t = \mathbf{y}_{t-1}, \forall t \in
 [1,...,m]$.

\subsection{Closed-form Solution}
Consider the case where the $d$ dimensional mapping function $\mathbf{y}=f(\mathbf{x})$ can be
any function for the temporal data.

Let the objective function be
\begin{equation}
J = E(||\mathbf{y}_t - \mathbf{y}_{t-1}||^2) - \log\det(cov(\mathbf{y}))
\end{equation}

Suppose the dynamics is a Markov chain on $n$ points ${\mathbf{y}_1, ..., \mathbf{y}_n}$, then
Eq. (2) can be rewritten as
\begin{equation}
J = \sum_{ij}p_{ij}||\mathbf{y}_j - \mathbf{y}_i||^2 - \log\det\Bigg(\sum_ip_i\mathbf{y}_i\mathbf{y}_i^T-\bigg(\sum_ip_i\mathbf{y}_i\bigg)\bigg(\sum_ip_i\mathbf{y}_i\bigg)^T\Bigg)
\end{equation}
where $p_i$ is the stationary distribution of the Markov chain and $p_{ij}$ is 
the stationary distribution of adjacent pairs.

Let $L$ be the Laplacian matrix defined as
\begin{equation}
L_{ij} = \left\{\begin{matrix}
p_i-p_{ii} \qquad \quad \textrm{if i==j}\\ 
-(p_{ij} + p_{ji})/2 \quad \textrm{otherwise}
\end{matrix}\right.
\end{equation}

Let $D$ be the diagonal matrix such that $D_{ii} = p_i$.

Eq. (2) has a closed-form solution for the internal representation $Y = [\mathbf{y}_i]$:

\begin{equation}
Y = U(2U^TLU)^{-1/2}R,
\end{equation}

where $U$ is the matrix whose column vectors are the generalized eigenvectors of
 $L$ and $D$, which is equivalent to the normalized-cut solution \cite{Shi00}, and $R$ is an arbitrary rotation matrix (i.e. $RR^T = I$). See Appendix A.1 for the proof.


\subsection{Online Learning Algorithm}
Although the above closed-form solution is appealing, it is computationally 
expensive for long video sequences and is not practical for real world 
situations where frames are captured one at a time. Therefore, we derive an 
online learning algorithm that can be applied to a deep neural network (See Appendix A.2 for the derivation).

Let $\mathbf{y}_t$ be the current output from the neural network, $\hat{\mathbf{y}}_t$ be the 
short-term moving average, and $\bar{\mathbf{y}}_t$ be the long-term moving average. 
Then $\hat{\mathbf{y}}_t$ and $\bar{\mathbf{y}}_t$ can be updated as follows:

\begin{align}
\hat{\mathbf{y}}_t &= (1 - \mu)\hat{\mathbf{y}}_{t-1} + \mu \mathbf{y}_t, \\
\bar{\mathbf{y}}_t &= (1 - \epsilon)\bar{\mathbf{y}}_{t-1} + \epsilon \mathbf{y}_t,
\end{align}

where $0 < \epsilon \ll \mu < 1$.

And short-term covariance $W_t$ and long-term covariance $B_t$ can be updated
as follows:

\begin{align}
W_t &= (1 - \mu)W_{t-1} + \mu (\mathbf{y}_t - \hat{\mathbf{y}}_t) (\mathbf{y}_t - \hat{\mathbf{y}}_t)^T, \\
B_t &= (1 - \epsilon)B_{t-1} + \epsilon (\hat{\mathbf{y}}_t - \bar{\mathbf{y}}_t) (\hat{\mathbf{y}}_t - \bar{\mathbf{y}}_t)^T.
\end{align}

The update rule for the parameters is:

\begin{equation}
\theta_t = \theta_{t-1} - \eta \frac{\partial J}{\partial \mathbf{y}_t} \frac{\partial \mathbf{y}_t}{\partial \theta},
\end{equation}
where
 \begin{equation}
\frac{\partial J}{\partial \mathbf{y}_t} = \big(\mathbf{y}_t - \hat{\mathbf{y}}_t\big)^TW_t^{-1} - \big(\hat{\mathbf{y}}_t - \bar{\mathbf{y}}_t\big)^TB_t^{-1}.
\end{equation}

\subsection{Two unsupervised learning layers}
We implemented the online learning algorithm with two unsupervised learning layers
 -- one for fully connected layers and the other for convolutional layers.
For the output from fully connected layers, the covariances are calculated with
respect to the whole vector; while for the output from convolutional layers,
the covariances are calculated with respect to each feature map channel.

These layers can be added to different deep learning models and at different
layers for providing a way to directly improve feature representation in each layer in an unsupervised learning manner. The weight term
$\mu$ in Eq. 3 is determined based on the distance between the unsupervised 
layer and the input layer. The closer it is to the input, the larger $\mu$
should be. This is because at lower level, the receptive field of a node is
smaller and the corresponding input signal tends to change faster. Therefore,
the proposed unsupervised learning layers can capture both fast and slow 
features at different depths of a DNN.

Unlike in regular DNN layers where all gradients are calculated at backward stages, 
in the unsupervised layers, the 
partial derivatives $\frac{\partial J_l}{\partial y_t}$ are calculated at 
forward stages and then combined with the gradient from the upper layers in the 
backward stage.
 
Algorithm~\ref{alg:online} shows a summary of the algorithm.

\begin{algorithm}[ht]
   \caption{Online Unsupervised Learning}
   \label{alg:online}
\begin{algorithmic}
   \STATE {\bfseries Input:} data $\mathbf{x}_t$, size $m$, $\mu$, $\epsilon$
   \STATE {\bfseries Output:} features $\mathbf{y}_t$ and parameters $\theta$
   \REPEAT
   \STATE Initialize $\hat{\mathbf{y}}_0$, $\bar{\mathbf{y}}_0$, $W_0$, $B_0$, $\theta_0$
   \FOR{$t=1$ {\bfseries to} $m$}
   \STATE forward regular layer: get output $\mathbf{y}_t = f_\theta(\mathbf{x}_t)$
   \STATE forward unsupervised layer: update $\hat{\mathbf{y}}_t$, $\bar{\mathbf{y}}_t$, $W_t$, $B_t$, $\frac{\partial J_l}{\partial \mathbf{y}_t}$ 
   \STATE backward unsupervised layer: update $\frac{\partial J}{\partial \mathbf{y}_t}$
   \STATE backward regular layer: update $\theta_t$ using Eq. (10).
   \ENDFOR
   \UNTIL{stopping criterion met}
\end{algorithmic}
\end{algorithm}

In summary, the proposed UL layers can play two roles: they can be
 the cost function layer for providing global training signal; meanwhile they
 can be added to any regular neural network layers for providing local training
signals and combined with the training signals backpropagated from upper layers
for extracting features at different changing paces. Therefore, the UL layers can be used in 
either pure unsupervised or semi-supervised settings with a supervised cost function layer
and several UL layers following the regular hidden layers.  

\section{Experiments}
The proposed algorithm can be applied to any neural networks. We conducted the 
following experiments to evaluate the performance of the algorithm.

\subsection{Synthetic sequence}
To evaluate the convergence speed and robustness of the proposed algorithm 
to noise, we generated synthetic video sequences consisting of a set of 28 
randomly selected 2D 
points rotating around its centroid sampled at one frame per 5 degrees for 
training as shown in Fig. 1(a). The network consists of one fully connected layer followed by an 
unsupervised layer. The input size is 56 (=28 * 2) and the output 2. 
The hyperparameters are: learning rate=0.01, momentum=0.9, weight decay rate=0.1, $\mu=0.5, \epsilon=0.001$. 
Fig. 1(b) and (c) visualize the two weight matrices learned in the fully 
connected layer which corresponds to the two major eigenvectors of the input 
sequence. 
Fig. 2(a) indicates that the output (each point corresponds to the output 
$y_t$ with respect to the input frame $x_t$) encodes the rotation angle of each frame in the sequence. This demonstrates that the neural network trained with our proposed algorithm can encode both input shape in the weights and motion in the output. We decoded the output using linear regression with respect to 
the sin function
of rotation angles and calculated the total absolute errors between the output and the ground truth. Fig. 2(b) illustrates 
that the algorithm can converge within 10 epochs for both training and test set.
To evaluate the noise sensitivity of the proposed algorithm, we tested the 
algorithm on five noise levels ranging from 0 to 50\%. Fig. 2(c) demonstrates 
that the proposed algorithm exhibits a smooth degradation as the noise level is 
increased from 0 to 40\%.

\begin{figure}[h]
\begin{center}
\centerline{\includegraphics[width=\columnwidth]{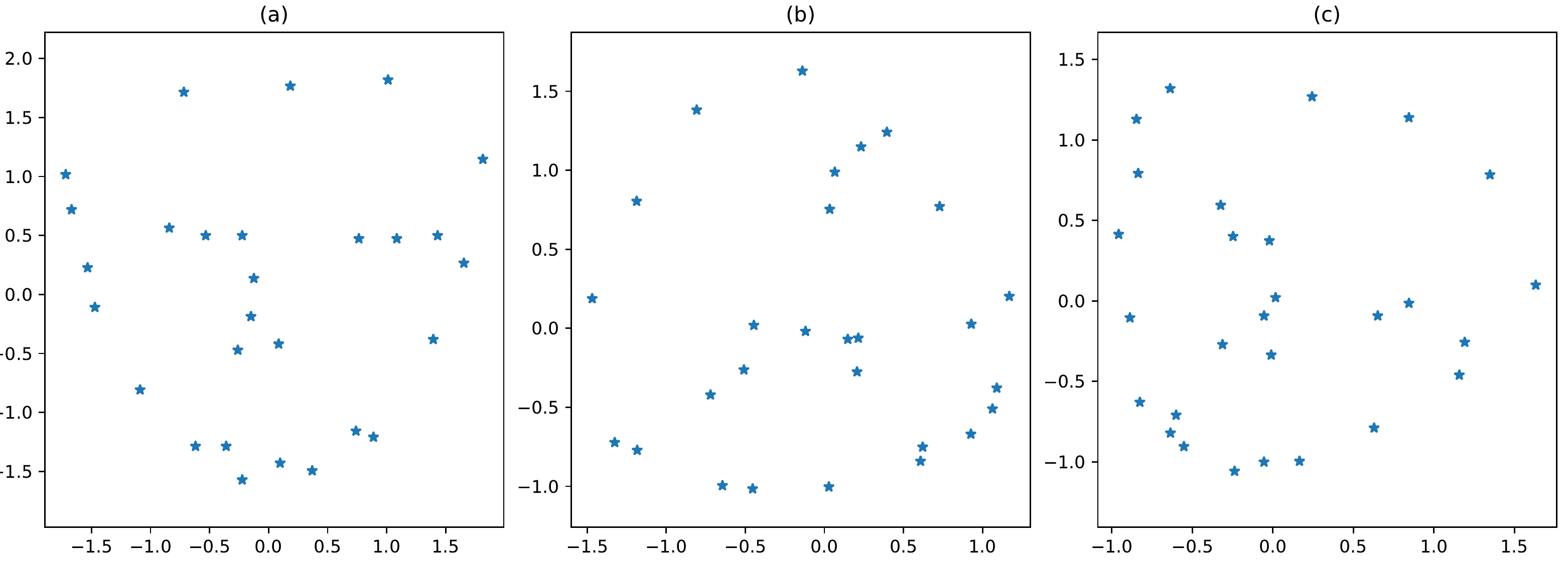}}
\caption{Synthetic sequences}
\label{syn}
\end{center}
\vskip -0.2in
\end{figure}

\begin{figure}[h]
\begin{center}
\centerline{\includegraphics[width=\columnwidth]{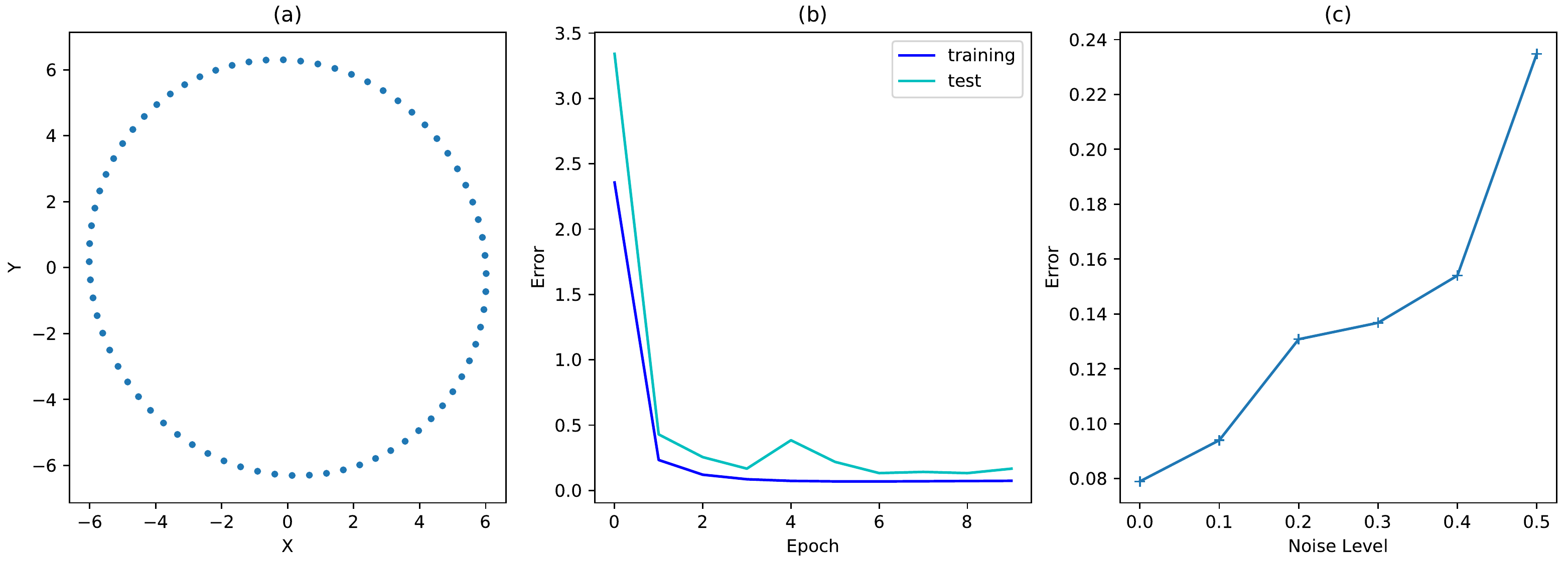}}
\caption{Experiments on synthetic sequences}
\label{syn2}
\end{center}
\vskip -0.2in
\end{figure}

\subsection{Head orientation estimation}
To gain an understanding of the representations learned in the convolutional 
layers, 
we trained a neural network consisting of a regular convolutional and a fully 
connected layers, each followed by an unsupervised layer for predicting the 
head orientation from a video sequence (generated from Basel Face Model
\footnote{\url{http://faces.cs.unibas.ch/bfm/?nav=1-0&id=basel_face_model}}.
The hyperparameters are: learning rate=1e-5, L2 regularization weight decay rate = 0.01 and they are set the same for the rest of experiments.

Fig. 3(a) shows one of the input image 
frame, Fig. 3(b) shows the learned four convolution kernels, and Fig. 3(c) shows the output from the convolutional layers which demonstrates that the learned 
representations captured the low-level features such as edges, corners and 
circles. 

\begin{figure}[h]
\begin{center}
\centerline{\includegraphics[width=\columnwidth]{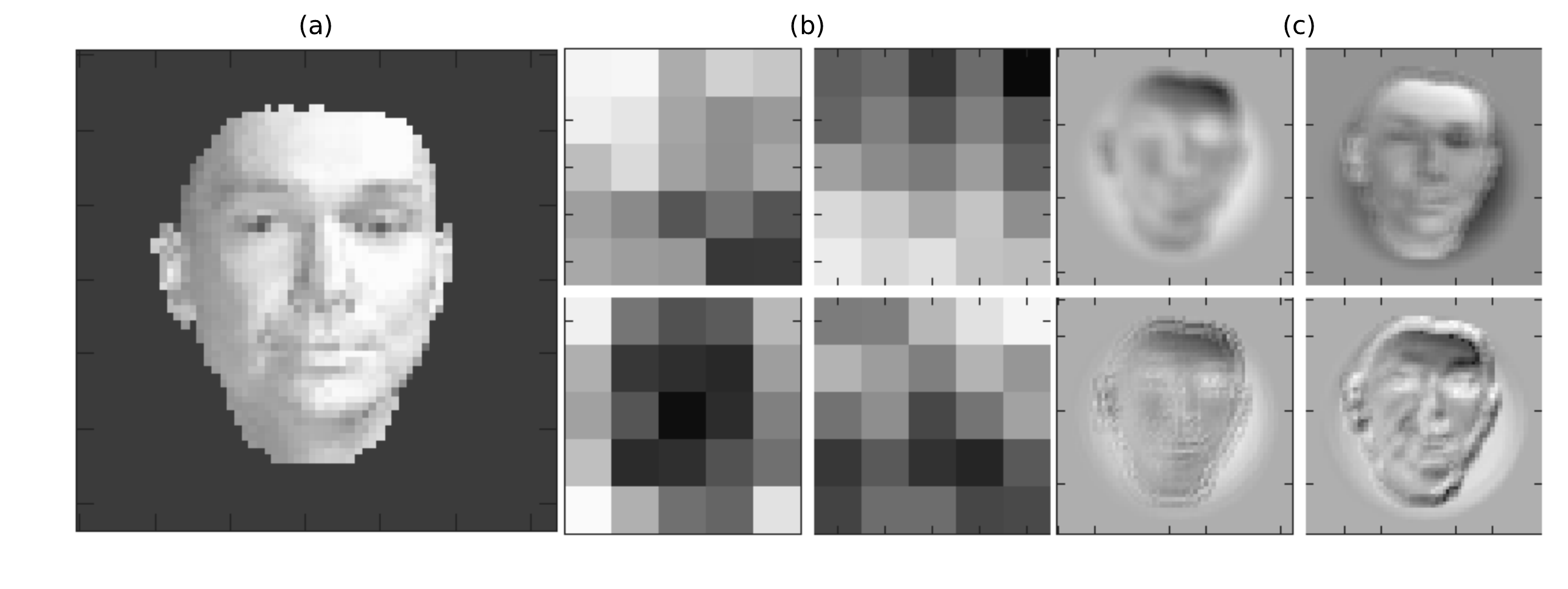}}
\caption{Head in-plane rotation sequence}
\label{head}
\end{center}
\vskip -0.2in
\end{figure}

We also trained the network for video sequences (generated from QMUL Multiview 
Face Dataset\footnote{\url{http://www.eecs.qmul.ac.uk/~sgg/QMUL_FaceDataset/}} consisting heads doing pan 
rotation as shown in Fig. 4. and evaluated the accuracy of the output from the 
network which encodes the sin of head orientation as shown in Fig. 5.

\begin{figure}[h]
\begin{center}
\centerline{\includegraphics[width=\columnwidth]{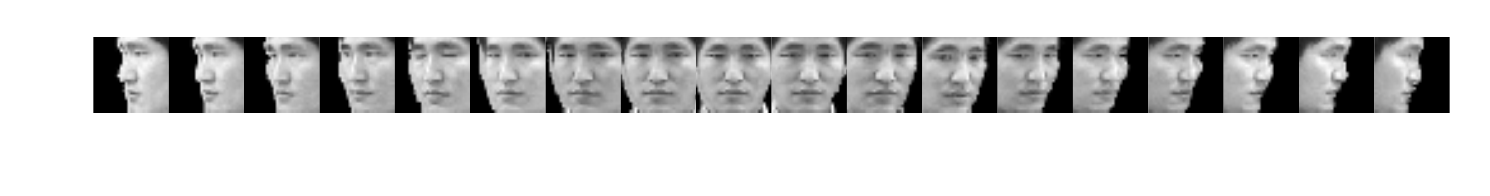}}
\caption{Head rotation sequence}
\label{head-pan}
\end{center}
\vskip -0.2in
\end{figure}

\begin{figure}[h]
\begin{center}
\centerline{\includegraphics[width=0.5\columnwidth]{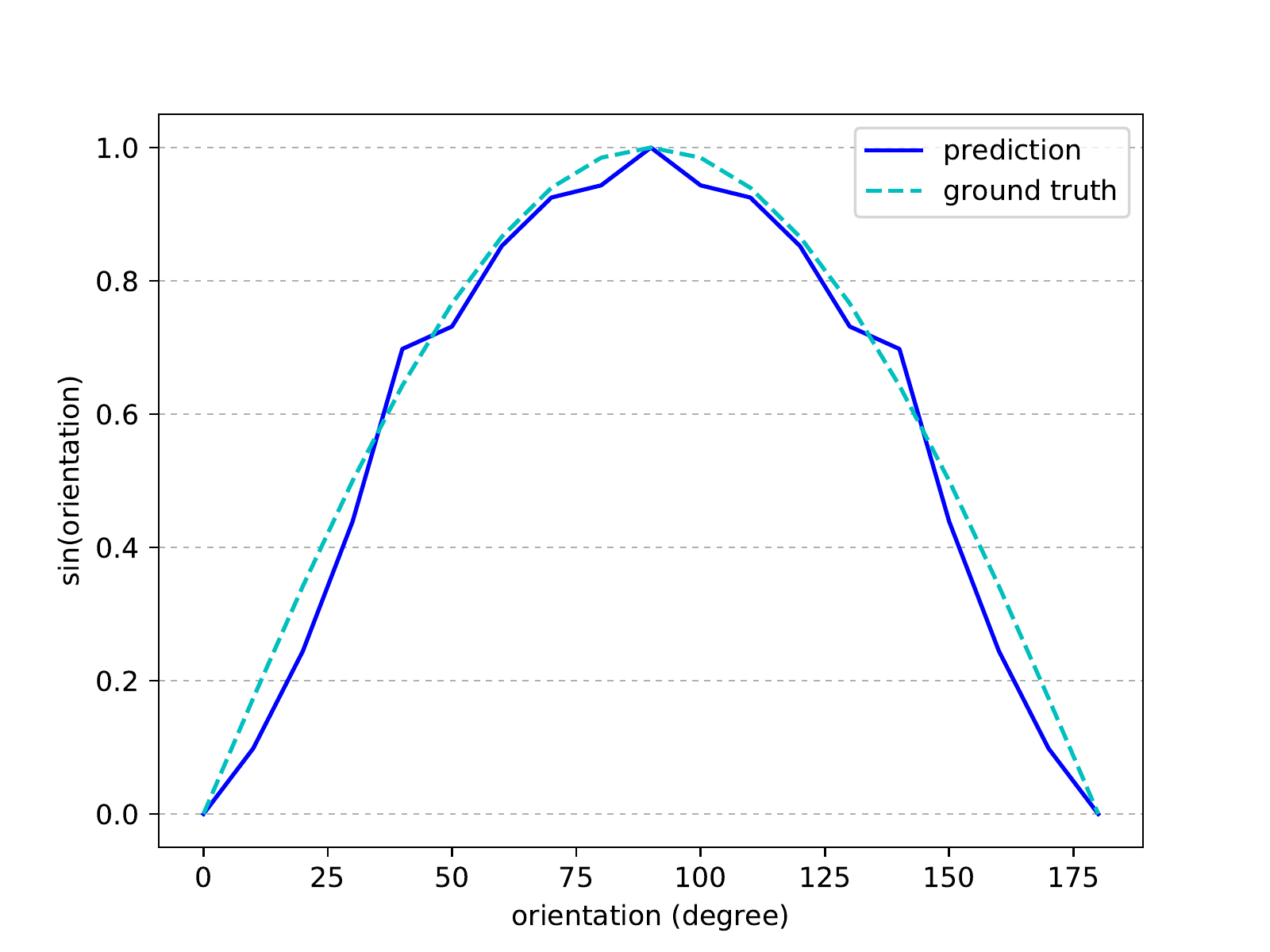}} 
\caption{Decoded head orientation}
\label{head-orientation}
\end{center}
\vskip -0.2in
\end{figure}

\subsection{Moving object localization}
To gain an understanding of the performance of the proposed 
algorithm on DNN with real videos, we evaluated the algorithm on two datasets 
as the same as in \cite{Stewart17}\footnote{\url{https://github.com/Russell91/labelfree}}. The first task is to predict the location of a pillow tossed by a 
person as shown in Fig. 6. The dataset consists of 65 sequences of a pillow in 
flight, totaling 602 images. Images are resized to 56 x 56 pixels. For 
comparison purpose, we employ the same NN architecture as in \cite{Stewart17} 
which consists of three
convolutional layers followed by two fully connected layers and one unsupervised
 learning layer.
Fig. 7(a) shows the qualitative results from our network after training on 32 sequences for 99 epochs. Since the author of \cite{Stewart17} did not provide ground truth for this dataset, we did not perform quantitative performance evaluation for this dataset. 

\begin{figure}[h]
\begin{center}
\centerline{\includegraphics[width=\columnwidth]{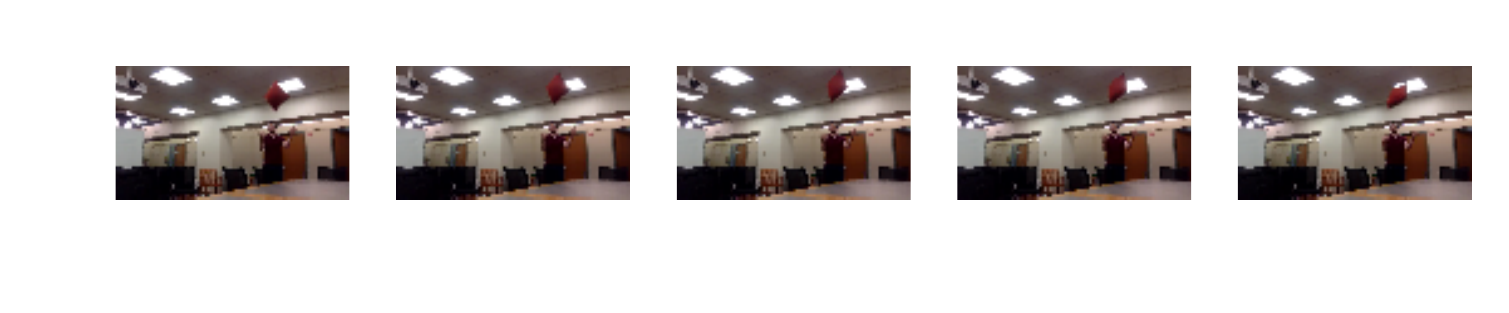}}
\caption{Pillow sequence}
\label{pillow}
\end{center}
\vskip -0.2in
\end{figure}

\begin{figure}[t!]
\centering

\subfigure[Pillow localization]{\label{pillow-loc}\includegraphics[width=0.4\columnwidth]{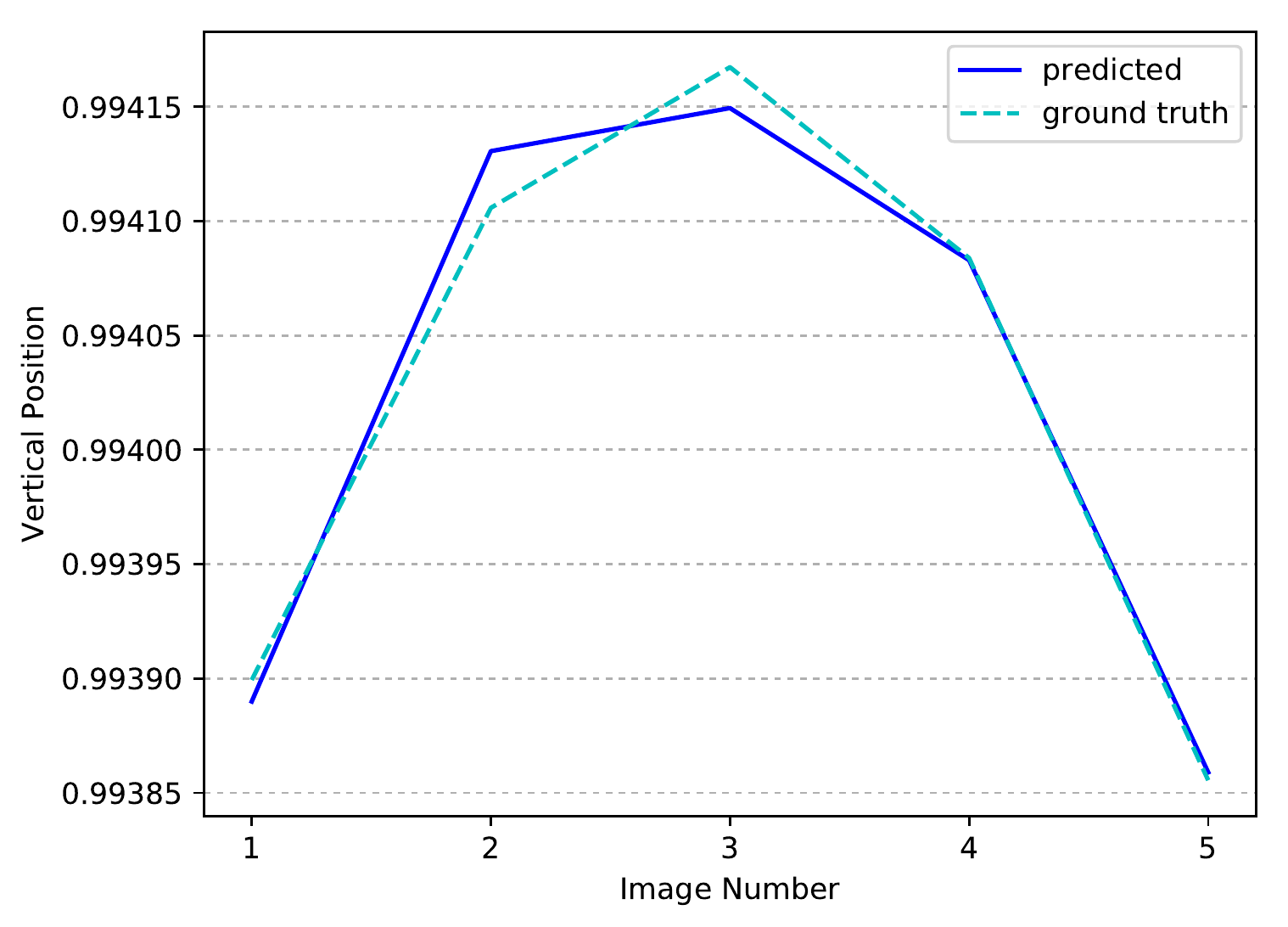}}
\subfigure[Walking man localization]{\label{loc-walk}\includegraphics[width=0.4\columnwidth]{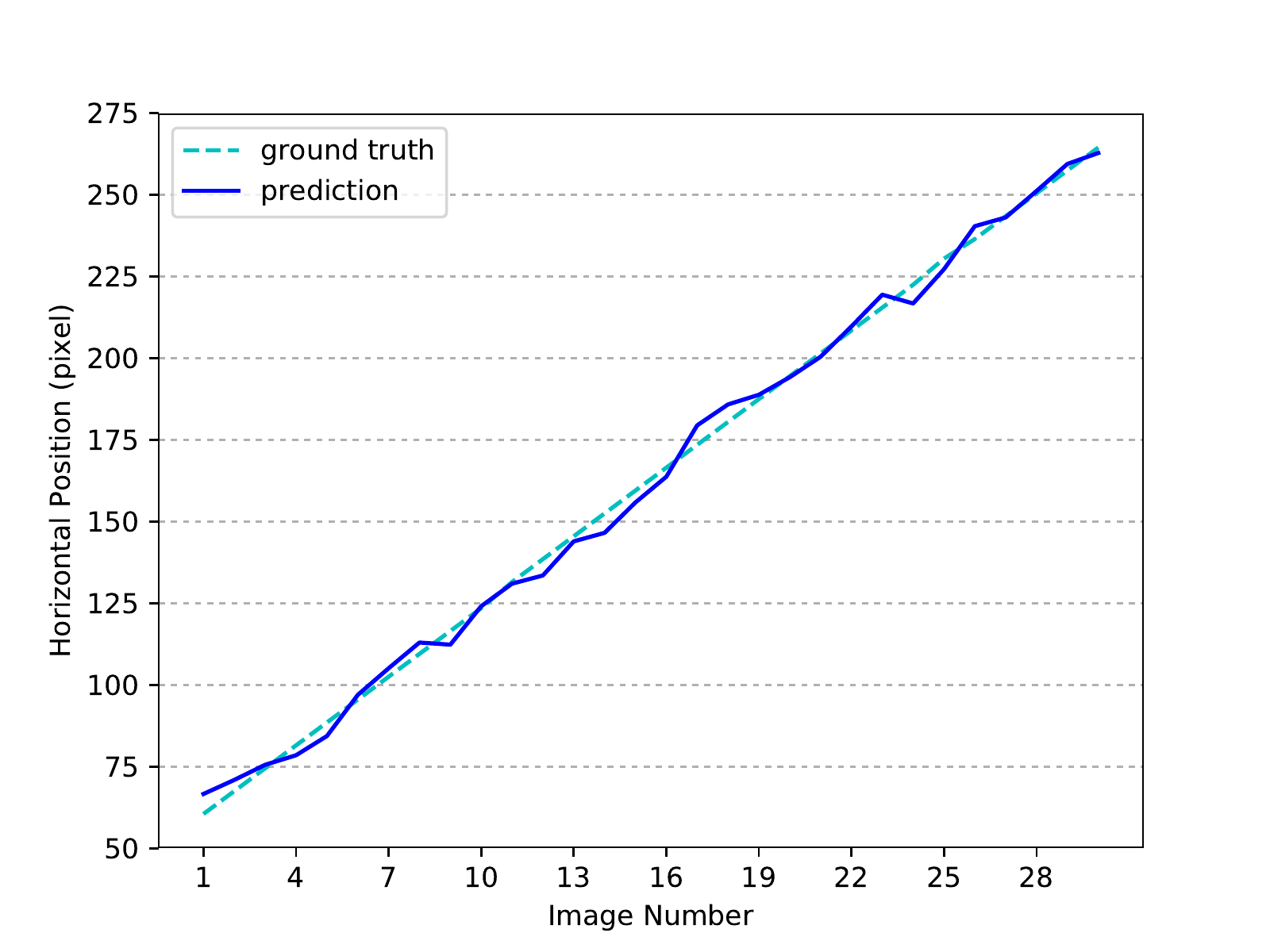}}
\caption{}
\end{figure}

\begin{figure}[h]
\begin{center}
\centerline{\includegraphics[width=\columnwidth]{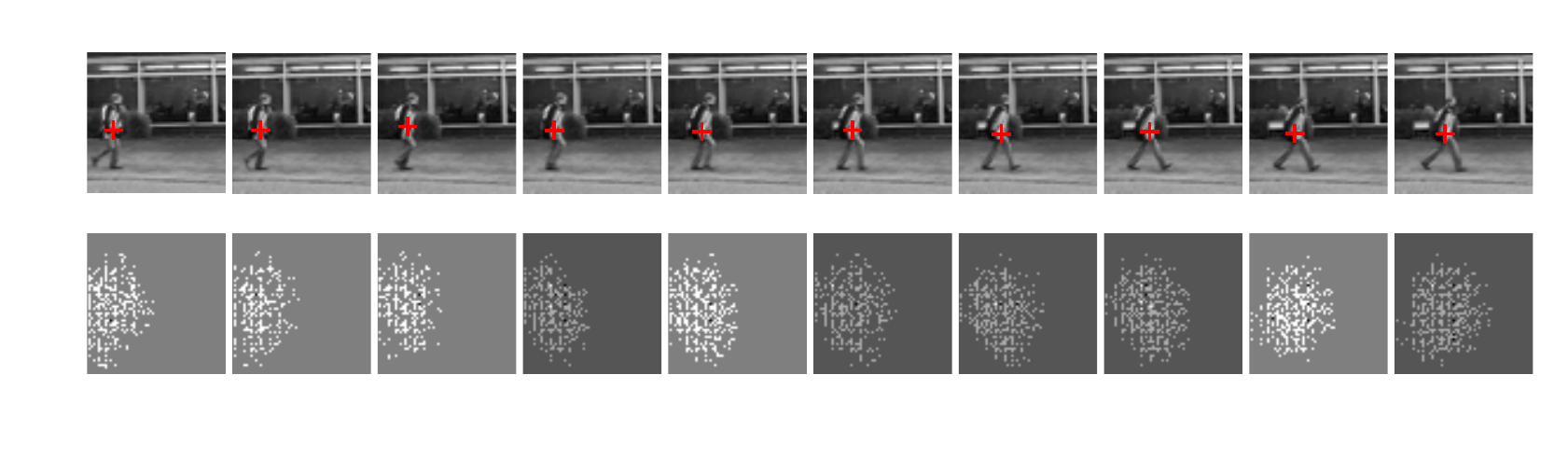}}
\caption{Moving object mask}
\label{mask}
\end{center}
\end{figure}

The second task is to predict the mask of a moving
object in a video as shown in Fig.~\ref{mask}. The dataset consisted of color images of
pixel size of (64, 64) with 10 sequences (261 frames) for training and 10 
sequences (246 frames) for validation. 

Fig. 8 shows the recovered masks and the centroids of the masks which locate the walking person correctly. Fig. 7(b) shows the 
comparison between the prediction and ground-truth. We estimate the correlation between our prediction and the ground truth which is 96.5\% and is better than the one in \cite{Stewart17}
which is 95.4\%. 

\section{Conclusion and future work}
We presents two unsupervised learning layers (UL layers) for label-free
 video analysis: one for fully connected layers, and the other for
convolutional layers. The proposed UL layers can play two roles: they can be
 the cost function layer for providing global training signal; meanwhile they
 can be added to any regular neural network layers for providing local training
signals and combined with the training signals backpropagated from upper layers
for extracting both slow and fast changing features at layers of
different depths.
Both a closed-form solution and an online learning algorithm implemented with
two UL layers are provided. Experiments with
 training on both unlabeled synthetic and natural videos 
demonstrated the effectiveness of the proposed algorithm and its potential 
applications to head orientation estimation, moving object localization.

Several parts of the presented approach could be tuned and studied in 
more detail. The proposed
unsupervised layers could be used in a semi-supervised setting. The temporal 
consistency constraint could be extended to spacial space. ... We leave these
questions for future research. We consider the present work as just a first 
step on the way to making an agent perform unsupervised learning in a real 
environment.
 
\section*{Acknowledgements}
We thank our IDL members Haonan Yu, Zhuoyuan Chen, Haichao Zhang, Tianbing Xu,
Yuanpeng Li, Peng Wang, Jianyu Wang, Xiaochen Lian and interns Qing Sun, Yaming Wang, 
Zhenheng Yang, Zihang Dai for their encouragement.


\bibliographystyle{plain}
\bibliography{nips_lz}

\appendix
\section{Appendix}
\subsection{Proof of the closed-form solution}
Let $\mathbf{p}$ be the vector $[p_i]$, then Eq. (3) can be rewritten as
\begin{equation}
J = 2tr(Y^TLY) - \log\det(Y^TDY - Y^T\mathbf{p}\mathbf{p}^TY)
\end{equation}

The minimum of $J$ is achieved when $\frac{dJ}{dY}=0$. That is
\begin{equation}
4LY - 2(D-\mathbf{p}\mathbf{p}^T)Y\sigma^{-1} = 0
\end{equation}
where $\sigma = Y^TDY - Y^T\mathbf{p}\mathbf{p}^TY$.

This shows that $Y$ is in a space spanned by $d$ generalized eigenvectors of
$L$ and $D-\mathbf{p}\mathbf{p}^T$. Let the $d$ eigenvectors and corresponding
eigenvalues be $\{\mathbf{u}_1, ..., \mathbf{u}_d\}$ and $\{\lambda_1, ..., \lambda_d\}$, respectively.
Then 
\begin{equation}
Y = UP,
\end{equation}
where $U=[\mathbf{u}_k]$ and $P=[p_{ij}]$.

And $\sigma=P^TU^T(D-\mathbf{pp}^T)UP = P^TP$.
Let $\Lambda = diag([\lambda_k])$, then Eq. (12) can be rewritten as 
\begin{align}
J &= 2tr(P^TU^T(D-\mathbf{pp}^T)U\Lambda P) - \log\det(P^TU^T(D-\mathbf{pp}^T)UP) \\
  &= 2tr(\sigma_u\Lambda PP^T) - \log\det(\sigma_u) \\
  &= 2tr(\Lambda A) - \log\det(A),
\end{align}
where $\sigma_u = U^T(D-\mathbf{pp}^T)U$, and $A = PP^T\sigma_u$.

$J$ is minimized when $A = (2\Lambda)^{-1}$, and $J = d + \log(\det(2\Lambda))$.
Correspondingly,
\begin{equation}
PP^T = (2U^TLU)^{-1}
\end{equation}

Therefore in order to minimize $J$, the $d$ eigenvalues should be the smallest
 $d$ ones except $0$.

Let $\mathbf{v}$ be a generalized eigenvector of $L$ and $D-\mathbf{pp}^T$ with
eigenvalue $\lambda \neq 0$. Note that vector $\mathbf{1}$ whose all elements i
are $1$ is also a generalized eigenvector of $L$ and $D-\mathbf{pp}^T$ with 
eigenvalue $0$, therefore $P^T\mathbf{v} = \mathbf{1}^TD\mathbf{v} = 0$. Hence
\begin{equation}
(D-\mathbf{pp}^T)\mathbf{v} = D\mathbf{v},
\end{equation} 
which means that $\mathbf{v}$ is also a generalized eigenvector of $L$ and $D$ and
 is equivalent to the normalized-cut solution \cite{Shi00}.

In summary, the closed-form solution for Eq. 1 is:
\begin{equation}
Y = UP = U(2U^TLU)^{-1/2}R,
\end{equation}
where $R$ is an arbitrary rotation matrix (i.e. $RR^T = I$).

\subsection{Derivation of the online learning algorithm}
We design the following cost function for the online learning algorithm:
\begin{equation}
J = \frac{1}{2}(\log\det(W) - \log\det(B))
\end{equation}
where $W$ and $B$ denote the short-term and long-term covariances of the output
$y_t$ from the neural network, respectively. The first term of Eq. 1 minimizes 
the entropy of short term distribution, while the second term maximizes the 
entropy of long term distribution. Therefore, the closer two input frames are,
the closer their corresponding output representations would be; and the farther
away the inputs are, the farther away their corresponding output representations
 should be. 

Let $\hat{\mathbf{y}}_t$ be the short-term moving average, and 
$\bar{\mathbf{y}}_t$ be the long-term moving average. We have
\begin{align}
W &= \frac{1}{N}\sum_i\sum_j(y_j - \hat{y}_i)(y_j - \hat{y}_i)^T \\
B &= \frac{1}{N}\sum_iN_i(\hat{y}_i - \bar{y})(\hat{y}_i - \bar{y})^T
\end{align}

and we can derive
\begin{align}
\frac{1}{2}\frac{\partial \log\det(W)}{\partial y_{n,p}}
  &= \frac{1}{2}\sum_{k,l}\frac{\partial \log\det(W)}{\partial W_{k,l}}\frac{\partial W_{k,l}}{\partial y_{n,p}} \\
  &= \frac{1}{2N}\sum_{k,l}(W^{-1})_{kl}(\delta_{k,p}y_{n,l} + \delta_{l,p}y_{n,k} - \delta_{k,p}\hat{y}_{i,k} \\
  &= \frac{1}{N}(W^{-1}(y_n - \hat{y}_i))_p
\end{align}
Similarly, we have
\begin{equation}
\frac{1}{2}\frac{\partial \log\det(B)}{\partial y_{n,p}} = \frac{1}{N}(B^{-1}(\hat{y}_n - \bar{y}))_p.
\end{equation}

Hence,
\begin{align}
\frac{1}{2}\frac{\partial \log\det(W)}{\partial \theta} &=  \frac{1}{N}\sum_i\sum_j(y_j-\hat{y}_i)^TW^{-1}\frac{\partial y_j}{\partial \theta} \\
\frac{1}{2}\frac{\partial \log\det(W)}{\partial \theta} &=  \frac{1}{N}\sum_i\sum_j(\hat{y}_j-\bar{y})^TB^{-1}\frac{\partial y_j}{\partial \theta}
\end{align}

In summary,
\begin{align}
\frac{\partial J}{\partial \theta} &= \frac{1}{2}\Bigg(\frac{\partial \log\det(W)}{\partial \theta} - \frac{\partial \log\det(W)}{\partial \theta}\Bigg) \\
 &= \frac{1}{N}\sum_i\sum_j((y_j - \hat{y}_i)^TW^{-1} - (\hat{y}_i - \bar{y})^TB^{-1})\frac{\partial y_j}{\partial \theta}
\end{align}
where $\hat{y}_i = \frac{1}{N_i}\sum_j y_j$ and $\bar{y} = \frac{1}{N}\sum_i y_i$.

\end{document}